\documentclass[runningheads]{llncs}
\usepackage{graphicx}
\usepackage{amsmath}
\usepackage{amssymb}
\usepackage{enumerate}
\usepackage{enumitem}
\usepackage{xcolor}
\usepackage{multirow}
\usepackage{tabu}
\usepackage{dashrule}
\usepackage{wrapfig}
\usepackage[colorlinks]{hyperref}

\usepackage{tikz}

\begin{document}
\title{TransBTS: Multimodal Brain Tumor Segmentation Using Transformer}
%
%
\author{
Wenxuan Wang$^{1}$, Chen Chen$^{2}$,
Meng Ding$^{3}$, Hong Yu$^{1}$, Sen Zha$^{1}$, Jiangyun Li$^{1,\dagger}$}
\authorrunning{Wenxuan Wang, Chen Chen,
Meng Ding, Hong Yu, Sen Zha, Jiangyun Li}
%
\institute{
School of Automation and Electrical Engineering, University of Science and Technology Beijing, China,
\email{s20200579@xs.ustb.edu.cn, g20198754@xs.ustb.edu.cn, g20198675@xs.ustb.edu.cn, leejy@ustb.edu.cn}\\
\and
Center for Research in Computer Vision, University of Central Florida, USA, \email{chen.chen@ucf.edu}\\
\and
Scoop Medical, Houston, TX, USA,
\email{meng.ding@okstate.edu}\\
$\dagger$ Corresponding author: Jiangyun Li
}

\maketitle              
\begin{abstract}

Transformer, which can benefit from global (long-range) information modeling using self-attention mechanisms, has been successful in natural language processing and 2D image classification recently. However, both local and global features are crucial for dense prediction tasks, especially for 3D medical image segmentation. In this paper, we for the first time exploit Transformer in 3D CNN for MRI Brain Tumor Segmentation and propose a novel network named TransBTS based on the encoder-decoder structure. To capture the local 3D context information, the encoder first utilizes 3D CNN to extract the volumetric spatial feature maps. Meanwhile, the feature maps are reformed elaborately for tokens that are fed into Transformer for global feature modeling. The decoder leverages the features embedded by Transformer and performs progressive upsampling to predict the detailed segmentation map. Extensive experimental results on both BraTS 2019 and 2020 datasets show that TransBTS achieves comparable or higher results than previous state-of-the-art 3D methods for brain tumor segmentation on 3D MRI scans. The source code is available at \url{https://github.com/Wenxuan-1119/TransBTS}.

\end{abstract}

\section{Introduction}
Gliomas are the most common malignant brain tumors with different levels of aggressiveness. 
Automated and accurate segmentation of these malignancies on magnetic resonance imaging (MRI) is of vital importance for clinical diagnosis. 

Convolutional Neural Networks (CNN) have achieved great success in various vision tasks such as classification, segmentation and object detection. Fully Convolutional Networks (FCN) \cite{fcn} realize end-to-end semantic segmentation for the first time with impressive results. U-Net \cite{unet} uses a symmetric encoder-decoder structure with skip-connections to improve detail retention, becoming the mainstream architecture for medical image segmentation.
Many U-Net variants such as U-Net++ \cite{unet++} and Res-UNet \cite{zhang2018road} further improve the performance for image segmentation. 
Although CNN-based methods have excellent representation ability, it is difficult to build an explicit \textbf{long-distance} dependence due to limited receptive fields of convolution kernels. This limitation of convolution operation raises challenges to learn global semantic information which is critical for dense prediction tasks like segmentation.

Inspired by the attention mechanism \cite{bahdanau2014neural} in natural language processing, existing research overcomes this limitation by fusing the attention mechanism with CNN models. Non-local neural networks \cite{wang2018non} design a plug-and-play non-local operator based on the self-attention mechanism, which can capture the long-distance dependence in the feature map but suffers from the
high memory and computation cost. Schlemper et al. \cite{schlemper2019attention} propose an attention gate model, which can be integrated into standard CNN models with minimal computational overhead while increasing the model sensitivity and prediction accuracy.
On the other hand, Transformer \cite{vaswani2017attention} is designed to model long-range dependencies in sequence-to-sequence tasks and capture the relations between arbitrary positions in the sequence. This architecture is proposed based \textit{solely on self-attention}, dispensing with convolutions entirely. Unlike previous CNN-based methods, Transformer is not only powerful in modeling global context, but also can achieve excellent results on downstream tasks in the case of large-scale pre-training. 

Recently, Transformer-based frameworks have also reached state-of-the-art performance on various computer vision tasks. Vision Transformer (ViT) \cite{vit} splits the image into patches and models the correlation between these patches as sequences with Transformer, achieving satisfactory results on image classification. DeiT \cite{deit} further introduces a knowledge distillation method for training Transformer.
DETR \cite{detr} treats object detection as a set prediction task with the help of Transformer. TransUNet \cite{chen2021transunet} is a concurrent work which employs ViT for medical image segmentation. We will elaborate the differences between our approach and TransUNet in Sec.~\ref{discuss}.

\textbf{Research Motivation.} The success of Transformer has been witnessed mostly on image classification. For dense prediction tasks such as segmentation, both local and global (or long-range) information is important. However, as pointed out by \cite{t2t}, local structures are ignored when directly splitting images into patches as tokens for Transformer. Moreover, for medical volumetric data (e.g. 3D MRI scans) which is \textbf{beyond 2D}, local feature modeling among continuous slices (i.e. depth dimension) is also critical for volumetric segmentation. 
We are therefore inspired to ask: \textit{How to design a neural network that can effectively model local and global features in spatial and depth dimensions of volumetric data by leveraging the highly expressive Transformer?}

In this paper, we present the \textit{first attempt} to exploit \textbf{Trans}former in 3D CNN for 3D MRI \textbf{B}rain \textbf{T}umor \textbf{S}egmentation (TransBTS). The proposed TransBTS builds upon the encoder-decoder structure.
The network encoder first utilizes 3D CNN to extract the volumetric spatial features and downsample the input 3D images at the same time, resulting in compact volumetric feature maps that effectively captures the local 3D context information. Then each volume is reshaped into a vector (i.e. token) and fed into Transformer for global feature modeling. The 3D CNN decoder takes the feature embedding from Transformer and performs progressive upsampling to predict the full resolution segmentation map.
Experiments on BraTS 2019 and 2020 datasets show that TransBTS achieves comparable or higher results than previous state-of-the-art 3D methods for brain tumor segmentation on 3D MRI scans. We also conduct comprehensive ablation study to shed light on architecture engineering of incorporating Transformer in 3D CNN to unleash the power of both architectures. We hope TransBTS can serve as a strong 3D baseline to facilitate future research on volumetric segmentation. 

\section{Method}
\subsection{Overall Architecture of TransBTS}
\begin{figure}
    \centering
    \includegraphics[width=0.99\textwidth]{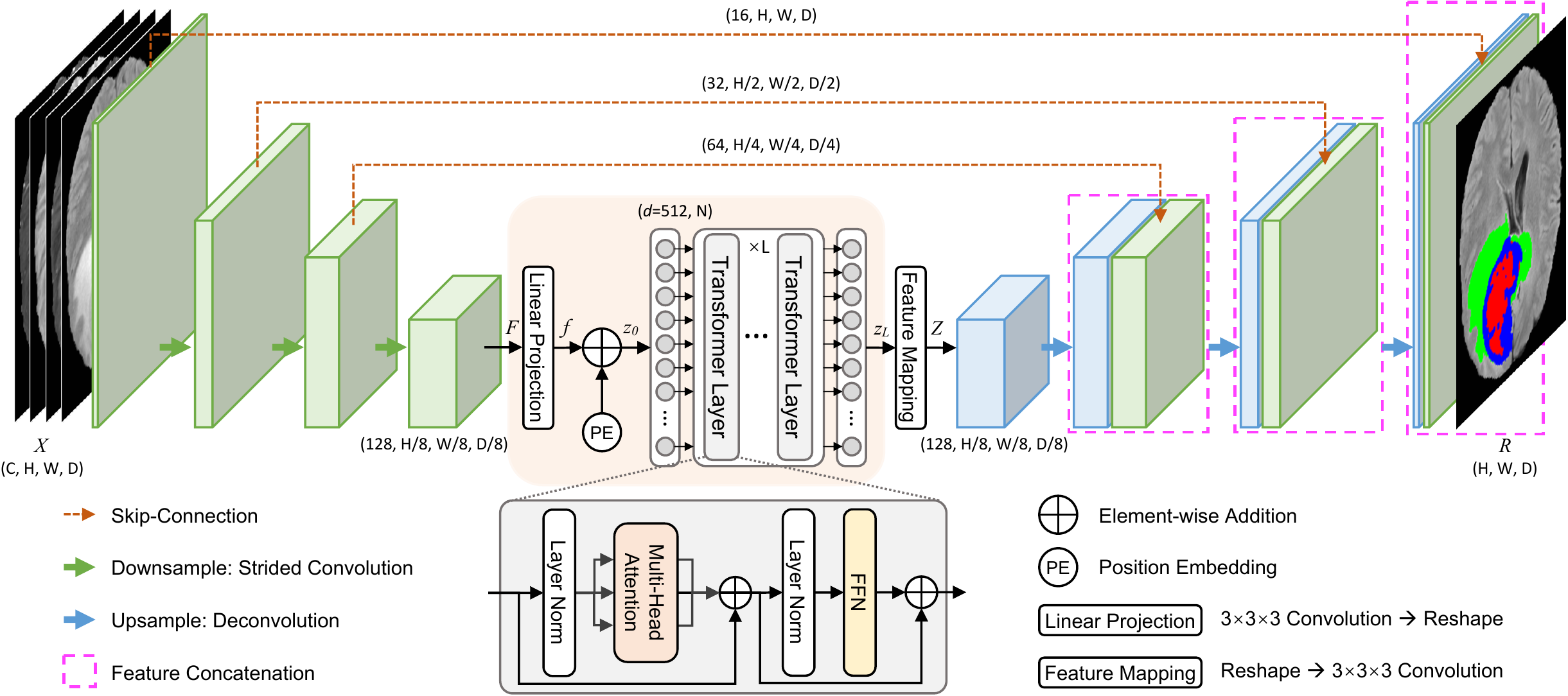}
    \caption{Overall architecture of the proposed TransBTS.}
    \label{fig1}
\end{figure}

An overview of the proposed TransBTS is presented in Fig. \ref{fig1}. 
Concretely, given an input MRI scan $X \in \mathbb{R}^{C \times H \times W \times D}$ with a spatial resolution of $H \times W$, depth dimension of $D$ (\# of slices) and $C$ channels (\# of modalities), we first utilize 3D CNN to generate compact feature maps capturing spatial and depth information, and then leverage the Transformer encoder to model the long distance dependency in a global space. After that, we repeatedly stack the upsampling and convolutional layers to gradually produce a high-resolution segmentation result. The network details of TransBTS are provided in the \textbf{Appendix}. Next, we will describe the components of TransBTS in detail.

\subsection{Network Encoder}
As the computational complexity of Transformer is quadratic with respect to the number of tokens (i.e. sequence length), directly flattening the input image to a sequence as the Transformer input is impractical. Therefore, ViT \cite{vit} splits an image into fixed-size ($16 \times 16$) patches and then reshapes each patch into a token, reducing the sequence length to $16^2$. For 3D volumetric data, the straightforward tokenization, following ViT, would be splitting the data into 3D patches. However, this simple strategy makes Transformer unable to model the image \textit{local context information across spatial and depth dimensions} for volumetric segmentation. 
To address this challenge, our solution is to stack the $3 \times 3 \times 3 $ convolution blocks with downsamping (strided convolution with stride=2) to gradually encode input images into low-resolution/high-level feature representation $F \in \mathbb{R}^{K \times \frac{H}{8} \times \frac{W}{8} \times \frac{D}{8}}$ ($K=128$), which is 1/8 of input dimensions of $H, W$ and $D$ ({overall stride (OS)=8}). In this way, rich local 3D context features are effectively embedded in $F$.
Then, $F$ is fed into the Transformer encoder to further learn long-range correlations with a global receptive field.

\noindent \textbf{Feature Embedding of Transformer Encoder.}
Given the feature map $F$, to ensure a comprehensive representation of each volume, a linear projection (a $3 \times 3 \times 3$ convolutional layer) is used to increase the channel dimension from $K=128$ to $d=512$. The Transformer layer expects a sequence as input. Therefore, we collapse the spatial and depth dimensions into one dimension, resulting in a $d \times N$ $(N=\frac{H}{8} \times \frac{W}{8} \times \frac{D}{8})$ feature map $f$, which can be also regarded as $N$ $d$-dimensional tokens. To encode the location information which is vital in segmentation task, we introduce the learnable position embeddings and fuse them with the feature map $f$ by direct addition, creating the feature embeddings as follows:
\begin{equation}
    z_{0}=f+PE=W \times F+PE
    \label{equation1}
\end{equation}
where $W$ is the linear projection operation, $PE \in \mathbb{R}^{d \times N}$ denotes the position embeddings, and $z_{0} \in \mathbb{R}^{d \times N}$ refers to the feature embeddings.

\noindent \textbf{Transformer Layers.}
The Transformer encoder is composed of $L$ Transformer layers, each of them has a standard architecture, which consists of a Multi-Head Attention (MHA) block and a Feed Forward Network (FFN). The output of the $\ell$-th ($\ell \in [1,2,...,L]$) Transformer layer can be calculated by:
\begin{equation}
    z_{\ell}^{'}=MHA(LN(z_{\ell-1}))+z_{\ell-1}
\end{equation}
\begin{equation}
    z_{\ell}=FFN(LN(z_{\ell}^{'}))+z_{\ell}^{'}
\end{equation}
where $LN(*)$ denotes the layer normalization and $z_{\ell}$ is the output of $\ell$-th Transformer layer. 

\subsection{Network Decoder}
In order to generate the segmentation results in the
original 3D image space ($H \times W \times D$), we introduce a 3D CNN decoder to perform feature upsampling and pixel-level segmentation (see the right part of Fig.~\ref{fig1}). 

\noindent \textbf{Feature Mapping.}
To fit the input dimension of 3D CNN decoder, we first design a feature mapping module to project the sequence data back to a standard 4D feature map. Specifically, the output sequence of Transformer $z_{L} \in \mathbb{R}^{d \times N}$ is first reshaped to $d \times \frac{H}{8} \times \frac{W}{8} \times \frac{D}{8}$. In order to reduce the computational complexity of decoder, a convolution block is employed to reduce the channel dimension from $d$ to $K$. Through these operations, the feature map $Z \in \mathbb{R}^{K \times \frac{H}{8} \times \frac{W}{8} \times \frac{D}{8}}$, which has the same dimension as $F$ in the feature encoding part, is obtained.

\noindent \textbf{Progressive Feature Upsampling.}
After the feature mapping, cascaded upsampling operations and convolution blocks are applied to $Z$ to gradually recover a full resolution segmentation result $ R \in \mathbb{R}^{H \times W \times D} $. Moreover, skip-connections are employed to fuse the encoder features with the decoder counterparts by concatenation for finer segmentation masks with richer spatial details.
 
\subsection{Discussion}
\label{discuss}
A very recent work TransUNet \cite{chen2021transunet} also employs Transformer for medical image segmentation. Here we want to highlight a few key distinctions between our TransBTS and TransUNet. (1) TransUNet is a 2D network that  processes each 3D medical image in a \textbf{slice-by-slice} manner. However, our TransBTS is based on 3D CNN and processes all the image slices at a time, allowing the exploitation of better representations of continuous information between slices. In other words, TransUNet only focuses on the spatial correlation between tokenized image patches, but our method can model the long-range dependencies in both slice/depth dimension and spatial dimension simultaneously for volumetric segmentation. (2) As TransUNet adopts the ViT structure, it relies on pre-trained ViT models on large-scale image datasets. In contrast, our TransBTS has a flexible network design and is trained from scratch on task-specific dataset without the dependence on pre-trained weights. 

\section{Experiments}
\textbf{Data and Evaluation Metric.}
The first 3D MRI dataset used in the experiments is provided by the Brain Tumor Segmentation (BraTS) 2019 challenge \cite{menze2014multimodal,bakas2017advancing,bakas2018identifying}. It contains 335 cases of patients for training and 125 cases for validation. Each sample is composed of four modalities of brain MRI scans, namely native T1-weighted (T1), post-contrast T1-weighted (T1ce), T2-weighted (T2) and Fluid Attenuated Inversion Recovery (FLAIR). Each modality has a volume of $240\times240\times155$ which has been aligned into the same space. 
The labels contain 4 classes: background (label 0), necrotic and non-enhancing tumor (label 1), peritumoral edema (label 2) and GD-enhancing tumor (label 4). 
The segmentation accuracy is measured by the Dice score and the Hausdorff distance (95\%) metrics for enhancing tumor region (ET, label 1), regions of the tumor core (TC, labels 1 and 4), and the whole tumor region (WT, labels 1,2 and 4).
The second 3D MRI dataset is provided by the Brain Tumor Segmentation Challenge (BraTS) 2020 \cite{menze2014multimodal,bakas2017advancing,bakas2018identifying}. It consists of 369 cases for training, 125 cases for validation and 166 cases for testing. 
Except for the number of samples in the dataset, the other information about these two datasets are the same.

\noindent \textbf{Implementation Details.}
The proposed TransBTS is implemented in Pytorch and trained with 8 NVIDIA Titan RTX GPUs (each has 24GB memory) for 8000 epochs from scratch using a batch size of 16. We adopt the Adam optimizer to train the model. The initial learning rate is set to 0.0004 with a poly learning rate strategy, in which the initial rate decays by each iteration with power 0.9. The following data augmentation techniques are applied: (1) random cropping the data from $240\times240\times155$ to $128\times128\times128$ voxels; (2) random mirror flipping across the axial, coronal and sagittal planes by a probability of 0.5; (3) random intensity shift between [-0.1, 0.1] and scale between [0.9, 1.1]. The softmax Dice loss is employed to train the network and $L2$ Norm is also applied for model regularization with a weight decay rate of $10^{-5}$. In the testing phase, we utilize Test Time Augmentation (TTA) to further improve the performance of our proposed TransBTS.

\begin{table}[htpb]
\scriptsize
    \centering
    \caption{Comparison on BraTS 2019 validation set. } \vspace{-5pt}
    \label{tab:comparison}
    {
    \begin{tabular}{l|c|c|c|c|c|c}

        \hline
        \multirow{2}{*}{Method} & \multicolumn{3}{c}{Dice Score (\%) $\uparrow$} & \multicolumn{3}{|c}{Hausdorff Dist. (mm) $\downarrow$} \\
        \cline{2-7}
        & \bfseries ET & \bfseries WT & \bfseries TC & \bfseries ET & \bfseries WT & \bfseries TC\\
        \hline
        3D U-Net \cite{3dunet}  & 70.86 & 87.38 & 72.48 & 5.062 & 9.432 & 8.719\\
        V-Net  \cite{vnet}    & 73.89 & 88.73 & 76.56 & 6.131 & 6.256 & 8.705\\
        KiU-Net \cite{valanarasu2020kiu}             & 73.21 & 87.60 & 73.92 & 6.323 & 8.942 & 9.893\\
        Attention U-Net  \cite{oktay2018attention}   & 75.96 & 88.81 & 77.20 & 5.202 & 7.756 & 8.258\\
        Wang et al. \cite{wang20193d}         & 73.70 &89.40 & 80.70 & 5.994 & 5.677 & 7.357\\
        Li et al. \cite{li2019multi}           & 77.10 & 88.60 & 81.30 & 6.033 & 6.232 & 7.409\\
        Frey et al. \cite{frey2019memory}      & 78.7 & 89.6 & 80.0 & 6.005 & 8.171 & 8.241\\
        Myronenko et al. \cite{myronenko2019robust}  & \textbf{80.0} & 89.4 & \textbf{83.4} & 3.921 & 5.89 & 6.562\\
        
        \hline
        \bf{TransBTS w/o TTA}  & 78.36 &  88.89 & 81.41 & 5.908 & 7.599 & 7.584\\
        \bf{TransBTS w/ TTA}   & 78.93 & \textbf{90.00} & 81.94 & \textbf{3.736} & \textbf{5.644} & \textbf{6.049}\\
        \hline
    \end{tabular}
    }
    
\end{table}

\begin{table}[htpb]
\scriptsize
    \centering
    \caption{Comparison on BraTS 2020 validation set.}
    \label{tab:comparison2020}
    {
    \begin{tabular}{l|c|c|c|c|c|c}

        \hline
        \multirow{2}{*}{Method} & \multicolumn{3}{c}{Dice Score (\%) $\uparrow$} & \multicolumn{3}{|c}{Hausdorff Dist. (mm) $\downarrow$} \\
        \cline{2-7}
        & \bfseries ET & \bfseries WT & \bfseries TC & \bfseries ET & \bfseries WT & \bfseries TC\\
        \hline
        3D U-Net    \cite{3dunet}        & 68.76 & 84.11 & 79.06 & 50.983 & 13.366 & 13.607\\
        Basic V-Net \cite{vnet}              & 61.79 & 84.63 & 75.26 & 47.702 & 20.407 & 12.175\\
        Deeper V-Net  \cite{vnet}              & 68.97 & 86.11 & 77.90 & 43.518 & 14.499 & 16.153\\
        Residual 3D U-Net   & 71.63 & 82.46 & 76.47 & 37.422 & 12.337 & 13.105\\
        \hline
        \bf{TransBTS w/o TTA}          & 78.50 & 89.00 & 81.36 & \textbf{16.716} & 6.469 & 10.468\\
        \bf{TransBTS w/ TTA}          & \textbf{78.73} & \textbf{90.09} & \textbf{81.73} & 17.947 & \textbf{4.964} & \textbf{9.769}\\
        \hline
    \end{tabular}
    }
\end{table}

\subsection{Main Results}
\begin{figure}[!tp]
    \centering
    \includegraphics[width=\textwidth]{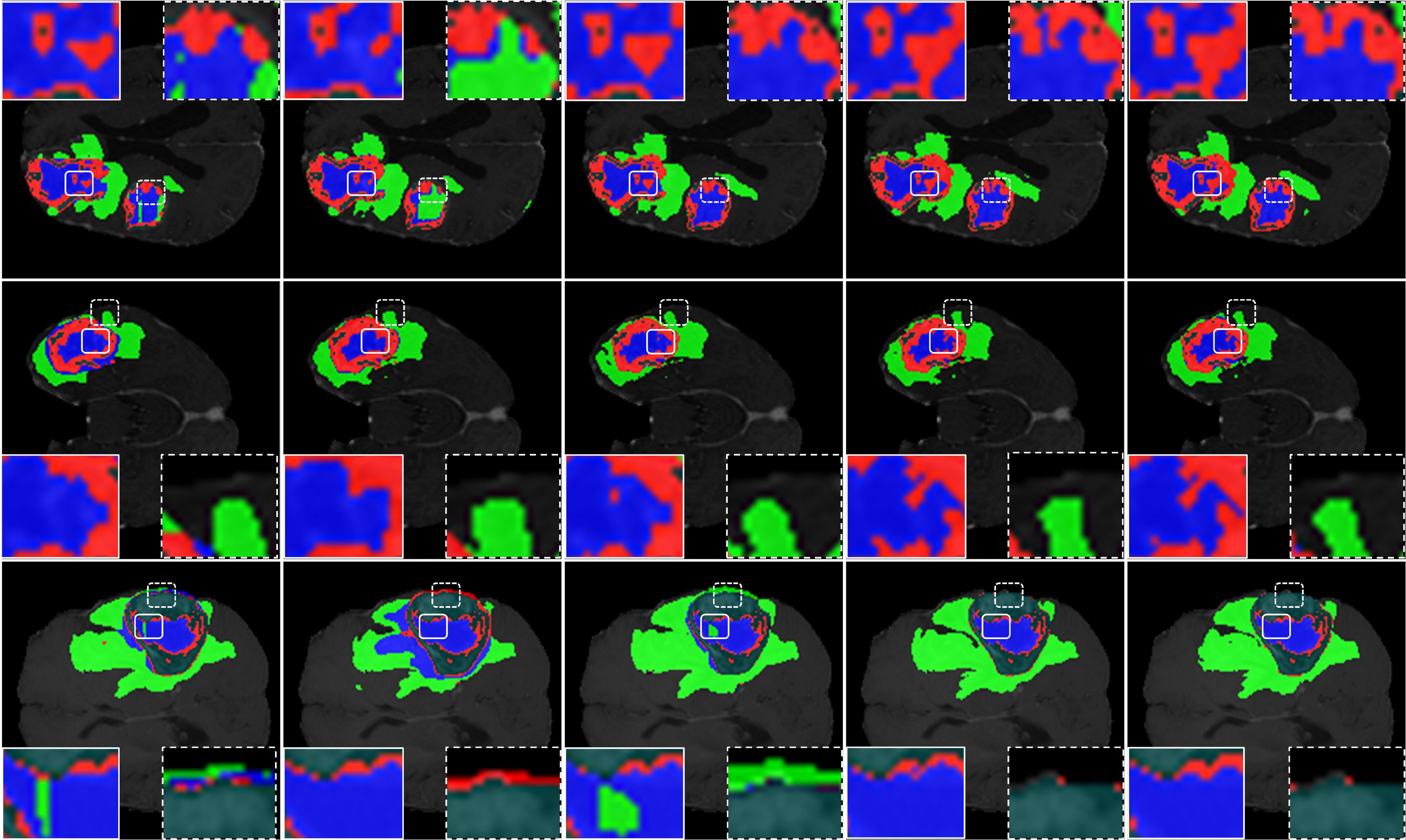}
    \begin{tabu} to \linewidth{X[1.0c] X[1.0c] X[1.0c] X[1.0c] X[1.0c]} \vspace{-8pt}
        \scriptsize{3D U-Net} & \vspace{-8pt} \scriptsize{VNet} & \vspace{-8pt} \scriptsize{Attention U-Net} & \vspace{-8pt} \scriptsize{TransBTS \textbf{(Ours)}} & \vspace{-8pt} \scriptsize{Ground Truth} \\
    \end{tabu}
    \caption{The visual comparison of MRI brain tumor segmentation results.  }
    \label{fig2}
\end{figure}

\textbf{BraTS 2019.} We first conduct five-fold cross-validation evaluation on the training set -- a conventional setting followed by many existing works. Our TransBTS achieves average Dice scores of $78.69\%$, $90.98\%$, $82.85\%$ respectively for ET, WT and TC. We also conduct experiments on the BraTS 2019 \textbf{validation} set and compare TransBTS with state-of-the-art (SOTA) 3D approaches. 
The quantitative results are presented in Table \ref{tab:comparison}.  TransBTS achieves the Dice scores of $78.93\%$, $90.00\%$, $81.94\%$ on ET, WT, TC, respectively,  which are comparable or higher results than previous SOTA 3D methods presented in Table \ref{tab:comparison}. In terms of Hausdorff distance metric, a considerable improvement has also been achieved for segmentation. Compared with 3D U-Net\cite{3dunet}, 
TransBTS shows great superiority in both metrics with significant improvements. This clearly reveals the benefit of leveraging Transformer for modeling the global relationships.
For \textbf{qualitative analysis}, we also show a visual comparison of the brain tumor segmentation results of various methods including 3D U-Net\cite{3dunet} , V-Net\cite{vnet}, Attention U-Net\cite{oktay2018attention} and our TransBTS in Fig.~\ref{fig2}. Since the ground truth for the validation set is not available, we conduct five-fold cross-validation evaluation on the training set for all methods. 
It is evident from Fig.~\ref{fig2} that TransBTS can describe brain tumors more accurately and generate much better segmentation masks by modeling long-range dependencies between each volume.

\noindent \textbf{BraTS 2020.} We also evaluate TransBTS on BraTS 2020 validation set and the results are reported in Table \ref{tab:comparison2020}. We directly adopt the hyperparameters on BraTS19 for model training, our TransBTS achieves Dice scores of $78.73\%$, $90.09\%$, $81.73\%$ and HD of 17.947mm, 4.964mm, 9.769mm on ET, WT, TC. Compared with 3D U-Net\cite{3dunet}, V-Net\cite{vnet} and Residual 3D U-Net, our TransBTS shows great superiority in both metrics with significant improvements. This clearly reveals the benefit of leveraging Transformer for modeling the global relationships.

\subsection{Model Complexity}
TransBTS has 32.99M parameters and 333G FLOPs which is a moderate size model. Besides, by reducing the number of stacked Transformer layers from 4 to 1 and halving the hidden dimension of the FFN, we reach a lightweight TransBTS which only has 15.14M parameters and 208G FLOPs while achieving Dice scores of $78.94\%$, $90.36\%$, $81.76\%$ and HD of 4.552mm, 6.004mm, 6.173mm on ET, WT, TC on BraTS2019 validation set. In other words, by reducing the layers in Transformer as a simple and straightforward way to reduce complexity ($54.11\%$ reduction in parameters and $37.54\%$ reduction in FLOPs of our lightweight TransBTS), the performance only drops marginally. Compared with 3D U-Net\cite{3dunet} which has 16.21M parameters and 1670G FLOPs, our lightweight TransBTS shows great superiority in terms of model complexity. Note that efficient Transformer variants can be used in our framework to replace the vanilla Transformer to further reduce the memory and computation complexity while maintaining the accuracy. But this is beyond the scope of this work.

\subsection{Ablation Study}
We conduct extensive ablation experiments to verify the effectiveness of TransBTS and justify the rationale of its design choices based on five-fold cross-validation evaluations on the BraTS 2019 training set.
(1) We investigate the impact of the sequence length ($N$) of tokens for Transformer, which is controlled by the overall stride (OS) of 3D CNN in the network encoder. (2) We explore Transformer at various model scales (i.e. depth ($L$) and embedding dimension ($d$)). (3) We also
analyze the impact of different positions of skip-connections. 

\noindent \textbf{Sequence length $N$.}
Table~\ref{table2} presents the ablation study of various sequence lengths for Transformer. The first row (OS=16) and the second row (OS=8) both reshape each volume of the feature map to a feature vector after downsampling. It is noticeable that increasing the length of tokens, by adjusting the OS from 16 to 8, leads to a significant improvement on performance. Specifically, $1.66\%$ and  $2.41\%$ have been attained for the Dice score of ET and WT respectively. Due to the memory constraint, after setting the OS to 4, we can not directly reshape each volume to a feature vector. So we make a slight modification to keep the sequence length to 4096, which is unfolding each $2\times2\times2$ patch into a feature vector before passing to the Transformer. We find that although the OS drops from 8 to 4, without the essential increase of sequence length, the performance does not improve or even gets worse. 

\begin{table}[!t]\scriptsize
    \newcommand{\tabincell}[2]{\begin{tabular}{@{}#1@{}}#2\end{tabular}}
	\centering
	\begin{minipage}{0.4\textwidth}
	\centering
	\caption{Ablation study on sequence length ($N$).} 
	\label{table2}
    \begin{tabular}{c|c|ccc}
    	\hline
            	\multirow{2}{*}{\textbf{OS}} &\multirow{2}{*}{\tabincell{c}{\textbf{Sequence} \\ \textbf{length}($N$)}} &\multicolumn{3}{c}{\textbf{Dice score(\%)}}  \\ \cline{3-5} & & ET & WT & TC \\

        \hline
        16 & 512 & 73.30 & 87.59 & \textbf{81.36} \\
        8 & 4096 & \textbf{74.96} & \textbf{90.00} & 79.96\\
        4 & 4096 & 74.86 & 87.10 & 77.46 \\
    	\hline
    \end{tabular}
    \end{minipage}
\hspace{1mm}
    \begin{minipage}{0.57\textwidth}
	\centering
	\caption{Ablation study on Transformer.} 
	\label{table3}
    \begin{tabular}{c|c|ccc}
    	\hline
            	\multirow{2}{*}{\textbf{Depth} ($L$)} &\multirow{2}{*}{\tabincell{c}{\textbf{Embedding} \textbf{dim} ($d$)}} &\multicolumn{3}{c}{\textbf{Dice score(\%)}}  \\ \cline{3-5} & & ET & WT & TC \\

        \hline
        4 & 384 & 68.95 & 83.31 & 66.89 \\
        4 & 512 & \textbf{73.72} & \textbf{88.02} & 73.14 \\
        4 & 768 & 69.38 & 83.54 & \textbf{74.16} \\
        1 & 512 & 70.11 & 85.84 & 70.95 \\
        8 & 512 & 66.48 & 79.16 & 67.22 \\
    	\hline
    \end{tabular}
    \end{minipage}

\end{table}


\noindent \textbf{Transformer Scale.}
Two hyper-parameters, the feature embedding dimension ($d$) and the number of Transformer layers (depth $L$), mainly determines the scale of Transformer. We conduct ablation study to verify the impact of Transformer scale on the segmentation performance.
For efficiency, we only train each model configuration for 1000 epochs. 
As shown in Table~\ref{table3}, the network with $d = 512$ and $L=4$ achieves the best scores of ET and WT. Increasing the embedding dimension ($d$) may not necessarily lead to improved performance ($L=4$, $d$: 512 vs. 768) yet brings extra computational cost. We also observe that $L=4$ is a ``sweet spot" for the Transformer in terms of performance and complexity.

\begin{table}

\scriptsize
	\centering
	\caption{Ablation study on the positions of skip-connections (SC).}
	\label{table4}
    \begin{tabular}{c|c|ccc}
    	\hline
            	\multirow{2}{*}{\textbf{Number of SC}}
            	&\multirow{2}{*}{\textbf{Position of SC}} &\multicolumn{3}{c}{\textbf{Dice score(\%)}}  \\ \cline{3-5} & & ET & WT & TC \\

        \hline
        3 & Transformer layer & 74.96 & 90.00 & 79.96 \\
        3 & 3D Conv (Fig.~\ref{fig1}) & \textbf{78.92} & \textbf{90.23} & \textbf{81.19} \\
    	\hline
    \end{tabular}
    
\end{table}

\noindent \textbf{Positions of Skip-connections (SC).}
To improve the representation ability of the model, we further investigate the positions for skip-connections (orange dash lines ``\begin{tikzpicture}[line width=1.2pt]
\draw [orange,dashed,->] (0,0)--(0.4,0);
\end{tikzpicture}" in Fig.~\ref{fig1}). The ablation results are listed in Table \ref{table4}.
If skip-connections are attached to the first three Transformer layers, it is more alike to feature aggregation from adjacent layers without the compensation for loss of spatial details. Following the traditional design of skip-connections from U-Net (i.e. attach to the 3D Conv layers as shown in Fig.~\ref{fig1}), considerable gains ($3.96\%$ and $1.23\%$) have been achieved for the important ET and TC, thanks to the recovery of low-level spatial detail information.

\section{Conclusion}
We present a novel segmentation framework that effectively incorporates Transformer in 3D CNN for multimodal brain tumor segmentation in MRI. The resulting architecture, TransBTS, not only inherits the advantage of 3D CNN for modeling local context information, but also leverages Transformer on learning global semantic correlations. Experimental results on two datasets (BraTS 2019 and 2020) validate the effectiveness of the proposed TransBTS. In future work, we will explore computational and memory efficient attention mechanisms in Transformer to develop efficiency-focused models for volumetric segmentation.

%
%
%
\bibliographystyle{splncs04}
\bibliography{reference}


\clearpage
\appendix

\section*{Appendix}

In this Appendix, we provide the network details of TransBTS in Table \ref{table5}.  

\begin{table}[]
\footnotesize
    \centering
        \caption{The design details of our propsoed TransBTS network. Conv3 denotes a $3 \times 3 \times 3$ convolutional layer; BN denotes Batch Normalization; DeConv denotes deconvolution layer. Each block of encoder and decoder is a residual block. Note that the size of input image is $ 4 \times 128 \times 128 \times 128 $.}
    \begin{tabular}{c|c|c|c}
        \hline
        stage & block name & details & output size \\
        \hline
        \multirow{12}*{3D CNN Encoder} & InitConv & Conv3, Dropout & $ 16 \times 128 \times 128 \times 128 $ \\
            \cline{2-4}
             ~ & EnBlock1 & $\left[ \begin{array}{ccc}
                  BN, & ReLU, & Conv3 \\
                  BN, & ReLU, & Conv3
             \end{array} \right]$ $\times 1$ & $ 16 \times 128 \times 128 \times 128 $ \\
             \cline{2-4}
             ~ & DownSample1 & Conv3(stride2) & $ 32 \times 64 \times 64 \times 64 $ \\
             \cline{2-4}
             ~ & EnBlock2 & $\left[ \begin{array}{ccc}
                  BN, & ReLU, & Conv3 \\
                  BN, & ReLU, & Conv3
             \end{array} \right]$ $\times 2$ & $ 32 \times 64 \times 64 \times 64 $ \\
             \cline{2-4}
             ~ & DownSample2 & Conv3(stride2) & $ 64 \times 32 \times 32 \times 32 $ \\
             \cline{2-4}
             ~ & EnBlock3 & $\left[ \begin{array}{ccc}
                  BN, & ReLU, & Conv3 \\
                  BN, & ReLU, & Conv3
             \end{array} \right]$ $\times 2$ & $ 64 \times 32 \times 32 \times 32 $ \\
             \cline{2-4}
             ~ & DownSample3 & Conv3(stride2) & $ 128 \times 16 \times 16 \times 16 $ \\
             \cline{2-4}
             ~ & EnBlock4 & $\left[ \begin{array}{ccc}
                  BN, & ReLU, & Conv3 \\
                  BN, & ReLU, & Conv3
             \end{array} \right]$ $\times 2$ & $ 128 \times 16 \times 16 \times 16 $ \\
             \hline
             \multirow{2}*{Transformer Encoder}  & Linear Projection & Conv3, Reshape & $ 512 \times 4096 $ ($d \times N$)\\
             \cline{2-4}
             ~ & Transformer & Transformer Layer $\times$ 4 & $ 512 \times 4096 $ ($d \times N$)\\
             \hline
        \multirow{13}*{3D CNN Decoder} & Feature Mapping & Reshape, Conv3 & $ 128 \times 16 \times 16 \times 16$ \\
            \cline{2-4}
            ~ & DeBlock1 &  $\left[ \begin{array}{ccc}
                  BN, & ReLU, & Conv3 \\
                  BN, & ReLU, & Conv3
             \end{array} \right]$  $\times 2$ & 
             $ 128 \times 16 \times 16 \times 16$ \\
            \cline{2-4}
            ~ & UpSample1 & Conv3, DeConv, Conv3 & $ 64 \times 32 \times 32 \times 32 $ \\
            \cline{2-4}
            ~ & DeBlock2 &  $\left[ \begin{array}{ccc}
                  BN, & ReLU, & Conv3 \\
                  BN, & ReLU, & Conv3
             \end{array} \right]$ $\times 1$ & $ 64 \times 32 \times 32 \times 32 $ \\
            \cline{2-4}
            ~ & UpSample2 & Conv3, DeConv, Conv3 & $ 32 \times 64 \times 64 \times 64 $ \\
            \cline{2-4}
            ~ & DeBlock3 &  $\left[ \begin{array}{ccc}
                  BN, & ReLU, & Conv3 \\
                  BN, & ReLU, & Conv3
             \end{array} \right]$  $\times 1$ & $ 32 \times 64 \times 64 \times 64 $ \\
            \cline{2-4}
            ~ & UpSample3 & Conv3, DeConv, Conv3 & $ 16 \times 128 \times 128 \times 128 $ \\
             \cline{2-4}
            ~ & DeBlock4 &  $\left[ \begin{array}{ccc}
                  BN, & ReLU, & Conv3 \\
                  BN, & ReLU, & Conv3
             \end{array} \right]$  $\times 1$ & $ 16 \times 128 \times 128 \times 128 $ \\
            \cline{2-4}
            ~ & EndConv & Conv1, Softmax & $ 4 \times 128 \times 128 \times 128 $ \\
            \hline
    \end{tabular}
    \label{table5}
\end{table}

\end{document}